\documentclass[runningheads]{llncs}
\usepackage[T1]{fontenc}

\usepackage[leftcaption]{sidecap}
\usepackage{latexsym}
\usepackage{amssymb}
\usepackage{amsmath}
\usepackage{booktabs}
\usepackage{enumitem}
\usepackage{graphicx}
\usepackage{color}
\usepackage{tikz}
\usetikzlibrary{arrows.meta, positioning}
\usepackage{multicol}
\usepackage{cancel}
\usepackage[normalem]{ulem}

\usepackage{mathptmx}
\usepackage{soul}\setuldepth{article}

\usepackage{xcolor}

\usepackage{tree-dvips}
\usepackage{qtree}

\usepackage{stmaryrd}
\usepackage{caption}
\usepackage{subcaption}

\newenvironment{varitemize}
{
\begin{list}{\labelitemi}
{\setlength{\itemsep}{0pt}
 \setlength{\topsep}{1.5pt}
 \setlength{\parsep}{0pt}
 \setlength{\partopsep}{0pt}
 \setlength{\leftmargin}{8.5 pt}
 \setlength{\rightmargin}{0pt}
 \setlength{\itemindent}{0pt}
 \setlength{\labelsep}{5pt}
 \setlength{\labelwidth}{10pt}
}}
{
 \end{list}
}

\newcommand{\BibTex}{B\kern-.05em{\sc i\kern-.025em b}\kern-.08em\Tex}

\newcommand{\defin}{=_{\textit{def}} }

\newcommand{\atm}{\mathit{Atm}}

\newcommand{\dec}{\mathit{Dec}}
\newcommand{\val}{\mathit{Val}}

\newcommand{\tagLabel}[2]{\tag{\textbf{#1}}\label{#2}}

\renewcommand{\phi}{\varphi}

\newcommand{\allins}{\Box}
\newcommand{\someins}{\diamond}

\newcommand{\takevalue}[1]{\mathsf{t}({#1})}

\newcommand{\putaway}[1]{}

\newcommand{\temporalbetter}{[\leq_{T}]}

\newcommand{\extemporalbetter}{\langle\leq_{T}\rangle}

\newcommand{\relevance}{\mathcal{R}}

\newcommand{\relHierarchy}{\mathcal{H}}
\newcommand{\relBinding}{\mathcal{B}}

\newcommand{\temporalorder}{\leq_{T}}
\newcommand{\stricttemporalorder}{<_{T}}

\newcommand{\relOp}{\mathtt{R^\forall}}
\newcommand{\relexOp}{\mathtt{R}^{\exists} }
\newcommand{\hOp}{\mathtt{H}}
\newcommand{\bOp}{\mathtt{B}}
\newcommand{\rOp}{\mathtt{R^\forall}}

\newcommand{\court}{\mathit{c}}

\newcommand{\Facts}{\mathtt{Facts}}
\newcommand{\Courts}{\mathtt{Courts}}
\newcommand{\Org}{\mathit{Jur}}
\newcommand{\Names}{\mathtt{Names}}

\newcommand{\cmtemp}{\mathbf{TJCM}}
\newcommand{\scmtemp}{\mathbf{\overline{TJCM}}}

\newcommand{\outcome}{\mathit{o}}

\newcommand{\logic}{TJCL}
\newcommand{\extlogic}{TJCL^{+}}

\newcommand{\exfuture}{\mathtt{\hat{F}}}

\newcommand{\expast}{\mathtt{\hat{P}}}

\newcommand{\pbinding}{\mathsf{PBinding}}
\newcommand{\poverruling}{\mathsf{POverruling}}

\newcommand{\name}{\mathit{n}}

\newcommand{\stricHierarchy}{\prec}
\newcommand{\eqHierarchy}{\cong}

\newcommand{\Lower}{\mathsf{Lower}}
\newcommand{\Higher}{\mathsf{Higher}}
\newcommand{\Samecourt}{\mathsf{SameCourt}}
\newcommand{\forallaccording}{\mathsf{According}^{\forall}}
 
\newcommand{\forallpoverruling}{\mathsf{POverruling}^{\forall}}
\usepackage{graphicx}
 \usepackage{orcidlink}
\begin{document}
\title{A Modal Logic for Temporal and Jurisdictional Classifier
Models}
%
%

%
\author{Cecilia Di Florio\inst{1,3}\orcidlink{0000-0002-8927-7414}\and
Huimin Dong\inst{2}\orcidlink{1111-2222-3333-4444} \and
Antonino Rotolo\inst{1}\orcidlink{2222--3333-4444-5555}}
\institute{{$^1$}University of Bologna, $^{2}$TU WIEN, $^{3}$University of Luxembourg}
\authorrunning{Di Florio, Dong and Rotolo}
%
%
%
\maketitle              
\begin{abstract}
Logic-based models can be used to build verification
tools for machine learning classifiers employed in the legal field.
ML classifiers predict the outcomes of new cases based on previous ones, thereby performing a form of case-based reasoning (CBR).
In this paper, we introduce a modal logic of classifiers designed to
formally capture legal CBR. We incorporate principles for resolving
conflicts between precedents, by introducing into the logic  the temporal dimension of cases and the hierarchy of courts within the legal system. 

\end{abstract}
\section{Introduction}

The use of machine learning (ML) classifiers to predict legal outcomes has been widely discussed in both academic literature and policy circles (see, e.g., \cite{Gan_Kuang_Yang_Wu_2021,Medvedeva2020-MEDUML,BexP21,ATKINSON2020103387}). In particular, judges have voiced specific concerns regarding the integration of ML into judicial decision-making, especially given that there is no guarantee that ML outcomes are normatively correct, accurate, or robust. To address these concerns, symbolic methods can be employed to formally verify the outcomes produced by judicial ML classifiers.

Especially within common law systems, one critical constraint that an ML classifier should satisfy is that of \emph{precedential constraint}, consistent with the doctrine of \emph{stare decisis}, which holds that prior judicial decisions constitute binding case law guiding future rulings. To develop verification methods that account for this constraint, we begin with an intuitive observation: ML classifiers derive the outcome of new cases based on a body of past cases, thereby performing a form of case-based reasoning (CBR). This observation is significant because the legal CBR literature already includes models of precedential constraint. Two well-known such models are the reason and result models proposed by Horty~\cite{Horty2011RR}. In fact, Liu and others~\cite{liu2022modelling} establish a correspondence between Horty’s models and classifier models (CMs) developed within the binary input classifier logic (BCL) framework~\cite{LiuLoriniJLC}. A key assumption in Horty's models is the \emph{consistency} of the initial case base: no prior case violated the precedential constraint~\cite{Horty2011RR}. As a consequence, the models preclude the possibility of retrieving conflicting precedents. 

Yet, the consistency requirement may not hold in real-world legal systems~\cite{Canavotto22}.
Conflicts of precedent do occur in legal practice, which is precisely why certain principles have been developed to resolve them. In common law systems, one of the most important is a \emph{temporal-hierarchical principle}: when precedents conflict, the more recent decision  issued by the higher court should prevail \cite{interpreting}. The importance of incorporating both temporal and hierarchical  elements into  models of precedential constraint, to better reflect actual legal reasoning, has already been emphasised in literature of AI and Law~\cite{Broughton,wyner,PrecedentsClash}. In particular, an extension of the semantics of CMs is proposed in \cite{PrecedentsClash}. 

In this work, we follow the idea that legal case-based reasoners are nothing but binary classifiers~\cite{Horty2011RR,liu2022modelling}. We propose a  modal logic for handling conflict of precedents. To this end, we extend BCL with both temporal and hierarchical operators. The resulting framework is theoretical, offering semantic and proof-foundations that may serve as a basis for future applications such as verification algorithms.

The paper is structured as follows. Section \ref{sec:conflict or precedents} introduces the notion of conflicting precedents. Section \ref{section:TJCM} extends BCL with the temporal and hierarchical elements.
Building on these elements, Section 4 provides a formal definition of both precedents and potentially binding precedents. A potentially binding precedent is a case that may constrain future decisions, unless it falls under some exceptions.
Finally, Section 5 formalises a temporal and hierarchical principle for resolving conflicts among precedents.

\section{Legal CBR and Conflict of Precedents}\label{sec:conflict or precedents}

Following the literature on legal CBR, our focus is on civil cases within common law systems, where judgments are rendered in favor of one of two opposing parties --- the plaintiff or the defendant. Suppose a ML classifier predicts the outcome of a case as favoring either the plaintiff or the defendant. The central question we want to address is: \emph{Is this outcome consistent with the precedential constraint?} To answer this question, we need to develop a suitable model.

Two well-known models of precedent constraint --- the \emph{reason} model and the \emph{result} model --- were introduced by Horty \cite{Horty2011RR}. Both represent cases using factors, which are legally relevant factual patterns supporting one party over the other. However within these models we cannot handle \emph{conflict of precedents}. To provide some intuition on the matter, in this section we focus on the result model. 
The result model captures \emph{a fortiori reasoning}: a case is forced to have a given outcome if there exists a precedent with the same outcome such that all differences between the two cases make the new case at least as strong in favor of that outcome as the precedent. Roughly,  a new case should have the same outcome as a precedent if it includes all the factors supporting the precedent’s outcome (or more), and no additional factors opposing it. Within the result model,  a \emph{case base} --- that is, a set of decided cases --- is \emph{consistent} if no case violates this \emph{a fortiori} constraint. In line with the CATO~\cite{Ashley1990ML} framework for CBR, we concentrate on  cases pertaining trade secret misappropriation and draw on a selected set of its factors:

\begin{varitemize}
\item Pro-plaintiff factors: $\mathsf{measure}$: ``The plaintiff had taken security measures to keep the secret'', $\mathsf{deceived}$: ``The defendant had obtained the secret by deceiving the plaintiff''; 
\item Pro-defendant factors: $\mathsf{reverse}$: ``The product is reverse-engineerable'', $\mathsf{disclosed}$: ``The plaintiff had voluntarily
disclosed the secret to outsiders''. 
\end{varitemize}
Now consider two cases, as summarized below:
\begin{varitemize}
\item Case 1: $\mathsf{\{reverse, disclosed, measure\}}$ --- decided for the plaintiff;
\item Case 2: $\mathsf{\{reverse, measure, deceived\}}$ --- decided for the defendant.
\end{varitemize}
The given case base violates the \emph{a fortiori} constraint: Case 2 contains fewer pro-defendant factors and more pro-plaintiff factors than Case 1, yet it is decided for the defendant, whereas Case 1 is decided for the plaintiff. The case base inconsistent.
Suppose  a ML classifier have to decide a new case:
\begin{varitemize}
\item Case 3: $\mathsf{\{reverse, measure, deceived, disclosed}\}$ --- outcome not known.
\end{varitemize}

Under the \emph{a fortiori} constraint, Case 3 should be decided for the plaintiff, based on Case 1, and for the defendant, based on Case 2. We have \emph{a conflict of precedents}. How should Case 3 be decided? Based solely on the available factual elements, the answer is uncertain. In this paper, we show that by taking into account other elements beside facts,  conflicts can be effectively managed. In particular, we will model a principle for resolving conflicts based on the temporal dimension of the cases --- when the cases were assessed --- and the hierarchical relationship between the courts that decided them.

In the example, we identified binding cases solely on the basis of factual comparison. That is, we considered as binding those cases that are applicable to the case at hand purely due to factual relevance. This relevance is grounded in \emph{a fortiori} reasoning.
Other criteria for determining factual relevance are adopted in different approaches.
In the strict model \cite{Broughton}, a precedent is considered applicable (and therefore binding) whenever its underlying reason applies to the new case. In many arguments-based approaches, the similarities and differences between cases can be strategically emphasized or downplayed \cite{Prakken2013}.
In all these approaches, factual relevance is regarded as sufficient for determining when a precedent is binding.
In this paper, we do not commit to any specific model of legal CBR. Our primary interest lies in addressing the problem of conflicting precedents. Henceforth, when we refer to a case as relevant to the case under consideration, we mean that it is factually applicable to the case at hand. Relevant cases, depending on specific criteria that we will introduce in the next section and formally define in Sections 4 and 5, may qualify as binding precedents.

\section{Language and  Temporal Jurisdictional Classifiers Models }\label{section:TJCM}

\subsection{Running Example}
\vspace{-1cm}
\begin{figure}[h]
\includegraphics[width=8.5 cm, height=4cm]{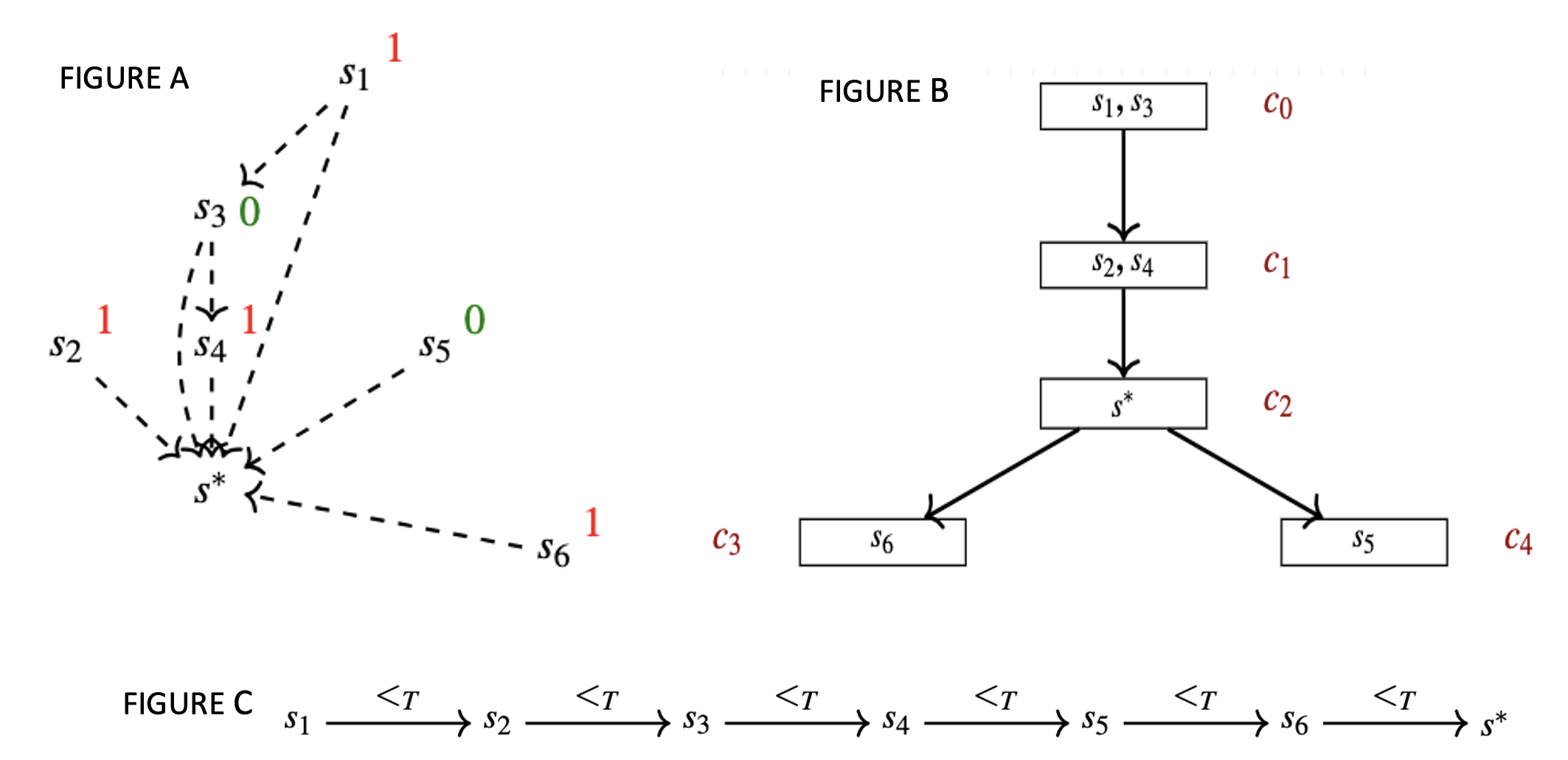}
\vspace{-1.5em}
\end{figure}

In the example from the previous section, we identified a conflict of precedents for Case 3: both Case 1 and Case 2 were relevant --- based on the \emph{a fortiori} constraint --- to Case 3, but they were decided in opposite directions.
From this point onward, we will consider a more complex and abstract example than the one presented above, in order to better illustrate the scenarios in legal practice that we aim to model in this paper. Let $s^{*}$ represent a new case to be decided based on previous cases $s_{1}, ..., s_{6}$. We denote a pro-defendant decision as $0$ and a pro-plaintiff decision as $1$. The relevance relations among the cases, as well as their decisions, are shown in Figure A.
All cases $s_i$ are all relevant to $s^{*}$, yet some have been decided as $0$ and others as $1$.  For example, both $s_3$ and $s_1$ are relevant for $s^{*}$, but $s_3$ is decided for the defendant, while $s_1$ is decided for the plaintiff --- we can think of $s^{*}$ as analogous to Case 3, with $s_1$ corresponding to Case 1 and $s_3$ to Case 2. 
Again, this seems to be a conflict among precedents. However, we overlooked three foundational elements of legal decision-making.

\noindent\textbf{Jurisdiction.} The precedent constraint is shaped by the legal system where cases are decided. A court is \emph{bound} only by decisions from courts with binding authority over it. Courts \emph{hierarchical and binding relationships}  vary depending on the jurisdiction.
\begin{varitemize}
\item[] We draw on the civil court system of England and Wales. At the top is the UK Supreme Court ($c_0$), which, is not formally bound by its own decisions. \footnote{Actually $c_0$  treat  its own decision as binding  when it sit in a partial panel~\cite{Allermuir}.} Next is the Court of Appeal ($c_1$), also self-bound, followed by the High Court ($c_2$), which we likewise assume binds itself. At the bottom are the County Courts (which are about 170~\cite{CrimeAndCourtsAct}), which do not issue binding decisions; we model only two of them ($c_3$ and $c_4$). Vertical \emph{stare decisis} applies:  higher courts bind lower ones, but not the reverse.
\end{varitemize}
Returning to our example, assume the cases are decided or to be decided by one of the four courts above. Specifically, there exists a hierarchical relationship between the cases, as illustrated in Figure B.
Consider cases $s_{5}$ and $s_{6}$, which are relevant to our case $s^{*}$. However, they were decided by courts $c_{3}$ and $c_{4}$, which do not hold binding authority over court $c_{2}$ deciding $s^{*}$. Therefore, \emph{$s_{6}$ and   $s_{5}$ are not binding for $s^{*}$}. We can eliminate $s_5$ and $s_6$ from binding precedents. But $s_1$, $s_2$, $s_3$, and $s_{4}$ may still be binding.

\noindent\textbf{Time.} 
Obviously, a case is constrained only by earlier decisions. Assume the timeline for our cases is as in Figure C. Then,  case $s^{*}$ is decided after $s_1$, $s_2$, $s_3$, and $s_{4}$, which could therefore be binding for  $s^{*}$. Time is also useful to model \emph{exceptions}.


\noindent\textbf{Exceptions.} The precedential constraint is subject to exceptions: there may be situations in which earlier cases lose their binding authority over subsequent ones. One example  is \emph{overruling}. 
Overruling occurs when a court rules against a relevant precedent \cite{interpreting,Rigoni_2014}. Not all courts can overrule. In line with common law systems, we assume a higher court  can always overrule a lower court’s decisions, and that a court  may overrule itself when it is not self-bound~\cite{interpreting}. 

For example, let us consider case $s_3$. This case was decided by the same court --- the Supreme Court $c_0$ --- \emph{after} the relevant precedent $s_1$. The decision in $s_3$ differs from that in $s_1$. Since we assumed that $c_0$ is not bound by its own prior decisions, it has the authority to overrule them. Accordingly, $s_1$ was overruled by $s_3$ and lost its binding authority over $s^*$. Therefore, we can  eliminate $s_1$ which is another source of conflict.  

Now, it seems that $s_4$, $s_3$, $s_2$ are left. What would be the decision for s$^*$? In this paper, we will formalise the notions introduced so far and attempt to answer this question. 


We conclude by mentioning that each case will be assigned a \emph{name}~\footnote{As in \cite{gorankoNames}, we use  term \emph{names} rather than  usual hybrid logic term \emph{nominals}.}. This is  consistent with legal practice, where every case has an identifiable name --- such as ``A vs B'' or a unique case identifier. Case naming will enables a hybrid approach in the logic.

\subsection{Temporal and Jurisdictional Classifier Models}

In this section we instantiate the hierarchical and temporal dimensions  in the framework of Classifier Models~\cite{LiuLoriniJLC}. 
First, we define the notion of jurisdiction, which characterises the structure of a specific legal system. This includes the identification of courts, the hierarchical relation among them  and which courts issue binding decision. 


\begin{definition}[Jurisdiction]\label{def:organisation}
A triple $(\Courts, \relHierarchy,\relBinding)$ is called a \emph{jurisdiction}, denoted as $\Org$, when: 
$\relHierarchy,\relBinding\subseteq \Courts\times\Courts$;  $\relHierarchy$ is transitive, irreflexive~\footnote{$\relHierarchy$ is irreflexive, when: for all  $\court\in\Courts$, it is {\bf not} the case that $\court \relHierarchy \court$.}.

\end{definition} \vspace{-0.2em}
\noindent
For $ \court_{i},  \court_{j}\in \Courts $, $ \court_{i} \relHierarchy \court_{j}$ is read as ``court $\court_{i}$ is hierarchically higher than  $\court_{j}$'';  $ \court_{i} \relBinding\court_{j}$ is read as ``$\court_{i}$ has binding authority on   $\court_{j}$.''
\vspace{-0.2em}

\begin{example}\label{Ex:Org}
The jurisdiction  in Figure B of  running example can be modeled as 
$\Org_{ex}= (\Courts_{ex},\relHierarchy_{ex},\relBinding_{ex})$, where: the set of courts is $\Courts_{ex} =\{\court_{0}, \court_{1}, \court_{2}, \court_{3}, \court_{4}\}$ and the hierarchy relation $\relHierarchy_{ex}\subseteq \Courts_{ex}\times\Courts_{ex}$ is $\relHierarchy_{ex} = \{(c_i,c_j)\mid i<j, 0\leq i\leq 2, 1\leq j\leq 4\}$. 
The binding relation is 
$\relBinding_{ex}=  \relHierarchy_{ex} \cup \{(\court_{1}, \court_{1}),(\court_{2}, \court_{2}) \}$.
\end{example} \vspace{-0.2em}


We introduce now the temporal and jurisdictional classifier models. We start by considering a set of possible input values of the classifiers, $\atm_{0}$. 
In $\atm_{0}$ there are: input variables describing the facts of the case, belonging to the countable set $\Facts$; the courts that can make the decision, belonging to the finite set $\Courts$; the name of each case belonging to the countable set $\Names$. Pertaining the $\Names$ we assume that there is a finite subset of names $\Names_{d}\subseteq \Names$ for decided cases. The sets $\Facts$, $\Courts$, and $\Names$ are non empty and pairwise disjoint to form $\atm_{0}$. In other words, $\atm_{0}= \Facts \cup \Names \cup \Courts$. The classifier outputs have values  in $\val = \{1, 0, ?\}$ where elements stand for \emph{plaintiff wins}, \emph{defendant wins} and \emph{absent  decision} respectively. For $\outcome \in \{0, 1\}$, the ``opposite'' $\overline{\outcome}$ is noted for the value $1 - \outcome$.  
\vspace{-0.2em}
\begin{definition}\label{def:classifiermodel}
    A temporal jurisdictional classifier model (TJCM) 
    is a tuple $C= (S, f, \Org, \temporalorder, \relevance)$ which satisfies these conditions: 
	\begin{varitemize}
	\item  $S\subseteq 2^{\atm_{0}} $ is a non-empty set s.t. $\forall s \in S \exists! \court \in \Courts:\court \in s$; 
	\item $f: S \longrightarrow \mathit{Val}$ is called a decision (or classification) function;
     \item  $\mid S_{d}\mid$ is finite, with $S_{d}=\{s\mid f(s)\neq ?\}$ be the set of decided states; 
        \item $ \Org$ is a jurisdiction;
        \item $\temporalorder$ is a total preorder on S~\footnote{$\temporalorder$ is a transitive and reflexive relation, with $\forall s, s'\in S$, either $s\temporalorder s'$ or $s'\temporalorder s$.}; 
        \item $\relevance\subseteq S\times S$ is a relation of relevant cases; 
        \item $\forall s \exists  n\in \Names: n\in s$; 
           \item $\forall s\in S\;\forall n\in \Names\cap s:  s\in S_{d} \Leftrightarrow  n\in \Names_{d}$; 
        \item  $\forall s, s'\in S : (s\neq s' \Rightarrow s\cap s'\cap \Names=\emptyset)$. 
     \end{varitemize}
\noindent The class of temporal and jurisdictional classifier
models is 
 $\cmtemp$ . 
\end{definition}



 $S$ is the set of states of the model, where each state contains a unique court. In this sense, a state $s \in S$ represents a case presented to a specific court $c\in\Courts$. 
The classification function maps each state to a value in the set $\{0, 1, ?\}$, representing the decision associated with the case. For each $s \in S$, we either have $f(s) = \outcome$, with $\outcome \in \{0, 1\}$ --- indicating that $s$ represents an assessed case --- or $f(s) = ?$, meaning $s$ is an unassessed or new case. The set $S_d$ of decided cases is finite. The structure $\Org$, as defined in Definition 1, specifies the hierarchical and binding relations among the courts. The relation $ \temporalorder$ is a temporal preorder, where $s \temporalorder s'$ is interpreted as “$s'$ was not assessed before $s$”. We say that $s$ is simultaneous to $s'$, denoted $s =_T s'$, if both $s\temporalorder s'$ and $s'\temporalorder s$. From $\leq_T$, we derive the strict temporal order $\stricttemporalorder$: $s \stricttemporalorder s'$ if and only if $s\temporalorder s'$ and $s' \not\leq_T s$. Thus, $s \stricttemporalorder s'$ means “$s$ was decided strictly before $s'$”.
 We write ${\leq_T}[s] = \{s' \mid s\temporalorder s'\}$ for the set of states that are assessed simultaneous or strictly later than $s$. By using a temporal \emph{preorder}, we allow two cases to be considered simultaneous. While it could be argued that, for any two cases, it is always possible to determine which was decided first, the temporal dimension is introduced here primarily to determine which precedents may be considered binding. We claim that, given a case $s$, a relevant case $s'$ decided on the same day or in the same week cannot be considered binding: the court assessing $s$ would not have had the opportunity to take $s'$ into account (e.g., because the reasons of $s'$ had not yet been published). Therefore, depending on the chosen temporal granularity, cases decided on the same day or in the same week may be considered simultaneous.
The binary relation $\relevance$ captures the notion of relevance: $s'\relevance s$ is read as “$s'$ is relevant for $s$”. We write $\relevance(s) = \{s' \mid s'\relevance s\}$ for the set of states relevant for $s$. The relevance relation is crucial in defining the notion of a precedent --- roughly, a precedent for a case is a relevant, previously assessed case. No specific definition of relevance is imposed in the model, nor are any particular properties (such as symmetry or transitivity) required for the relevance relation. This flexibility allows the model to accommodate various notions of relevance. 
Each state is associated with a name, and no two states share the same name. \footnote{As it happens in hybrid logic \cite{Blackburn_Rijke_Venema_2001}, a state  can have multiple names. This is not problematic: legal cases may have a common name like "A vs B" alongside a unique identifier. } All and only the decided state have names in the finite subset of names $Names_{d}$. We will say that state $s$ is $n$-named if $n\in s$.

For notational convenience, we extend the  relationships between  courts to states: $s \stricHierarchy s'$  iff $\relHierarchy(c',c)$ where $c\in s\cap \Courts$, $c'\in s'\cap \Courts$; $s \eqHierarchy s'$  iff  $c\in s\cap s'\cap \Courts$.


We have not  specified the relationship between hierarchical and binding relations, allowing our framework to accommodate different legal systems.   However, in this work, we  focus  on \emph{common law systems}. Within such systems, typically, 1) \emph{vertical stare decisis} applies, i.e. decisions of higher courts bind lower courts:
$ \text{ if } c_i\relHierarchy c_j \text{ then } c_i \relBinding c_j$;
2) lower courts have no binding power on higher courts and  two  non-hierarchically-related courts don't issue binding decisions one for the other: $ \text{ if not } c_i\relHierarchy c_j \text{ then not } c_i \relBinding c_j$; 3) \emph{horizontal stare decisis} may apply, i.e. some $c_i$ may be self-bound, namely it may be $c_i \relBinding c_i$. 
Ultimately from 1), 2) and 3) it follows\vspace{-4pt}
\begin{align*}\tag{SD}\label{eq:SD}
 \relHierarchy\subseteq \relBinding\subseteq \relHierarchy\cup \mathcal{I}  \\[-20pt]
\end{align*}
where $\mathcal{I}= \{(c,c) \mid c \in {\sf Courts}\}$ is the identity relation to identify the self-bound courts. 


Henceforth, we impose a condition on the temporal dimension of states of classifier models:
the new and undecided cases are strictly after  already assessed cases. 
\vspace{-4pt}
\begin{align*}\tag{O}\label{eq:T}
 \forall s,s'\in S : (f(s)\neq ? \,\&\,f(s')=?) \Rightarrow s<_{T}s'\\[-18pt]
 \end{align*}


 From now on we focus on $\scmtemp$, the subclass of TJCMs satisfying  $(\ref{eq:T})$, and ($\ref{eq:SD}$).

\begin{example}
We model the running example through TJCMs. Fix $\atm_{0}$, assuming the finite set of decided names is $\Names_{d}=\{\name_{1},..., \name_{6}\}$. Let $S_{ex}=\{s_{1},...,s_{6}, s^{*}\}\subseteq 2^{\atm_{0}}$. Pertaining names, assume  $n_{i}\in s_{i}$ and $s^{*}\cap \Names =\{n^{*}\}$. 
 $C_{ex}=(S_{ex},f_{ex}, \Org_{ex}, \temporalorder^{ex}, \relevance^{ex} )$ is the classifier model, where  $f_{ex}(s^{*})=?$, $f_{ex}(s_{1})= f_{ex}(s_{2})= f_{ex}(s_{4})=f_{ex}(s_{6})=1$, $f_{ex}(s_{3})=f_{ex}(s_{5})=0$; $\Org_{ex}$ is defined  in Ex.~\ref{Ex:Org}; $\relevance^{ex}(s^{*})= \{s_{1}, s_{2}, s_{3}, s_{4}, s_{5}, s_{6}\}$, $\relevance^{ex}(s_{3})= \{s_{1}\}$,  $\relevance^{ex}(s_{4})= \{s_{3}\}$; $s_{1}\stricttemporalorder s_{2}<_{T}\dots\stricttemporalorder s_{6}\stricttemporalorder s^{*}$. $C_{ex}$ satisfies $(\ref{eq:T})$,  ($\ref{eq:SD}$). 
\end{example}

\subsection{Temporal and Jurisdictional Classifers Logic}

Now we introduce the temporal and jurisdictional classifiers logic $\logic$. First, we will introduce the language to express the legal decision process in which the temporal and jurisdictional constraints are considered. For any $\outcome\in \val$, we call $\takevalue{\outcome}$ a decision atom, to be read as “the actual decision (or output) takes value $\outcome$". $\dec = \{\takevalue{\outcome} : \outcome\in \val$\} is the set of decision atoms. $\atm= \atm_{0}\cup \dec$.

\begin{definition}[Language]   
The modal language $\mathcal{L}(\mathit{Atm})$ is hence defined as: \vspace{-4pt}
\[
\varphi::= p \mid  \takevalue{\outcome} \mid  \hOp (c,c)\mid \bOp (c,c) \mid 
  \neg\varphi \mid \varphi \wedge\varphi \mid \allins \varphi  \mid \temporalbetter \varphi\mid \relOp\varphi
\]\\[-30pt]

where $p\in \atm_{0}$, $\takevalue{o}\in \dec$, and $\court\in \Courts$. 

\end{definition}
\noindent 
Let $n\in \Names $. We define the formula  $@_{n}\varphi$ as $ \allins(n \rightarrow \varphi )$. The formula has a hybrid logic flavour but is expressed in our language without explicit hybrid operators.

$\hOp (c_i,c_j)$ is read as ``The court $c_i$ is hierarchically higher than the court $c_j$''. $\bOp (c_i,c_j)$ is read as ``The court $c_i$ issues binding decisions for the court $c_j$''.
 $\allins \varphi$ is read  as ``$\varphi$ is universally true''.  
$\temporalbetter \varphi$ is read as ``$\varphi$ is the case and always will be the case''. Finally,  $\relOp\varphi$, is  read “$\varphi$ is true at all states that are relevant for the current one”.  
We define the dual operators as usual: $\diamond \varphi \defin \neg \allins \neg \varphi$, $\mathtt{R}^{\exists} \varphi \defin \neg \relOp \neg \varphi$,$\extemporalbetter\varphi\defin\neg\temporalbetter\neg \varphi$. Formulas of $\mathcal{L}(\mathit{Atm})$ are interpreted with respect to classifier models as follows.






\begin{definition}[Satisfaction relation]\label{truthcondCM}
	Let $C= (S, f, \Org, \temporalorder, \relevance)$ be a temporal jurisdictional classifier model. Let 
    $s \in S$. Then:\\[2pt]
\begin{tabular}{p{5cm}p{7.5cm}}
   $(C, s) \models p  \Longleftrightarrow p \in s$;  & $(C, s) \models \varphi\wedge\psi \Longleftrightarrow (C, s) \models \varphi \text{ and } (C, s) \models  \psi $; \\
   $(C, s) \models \mathsf{t} (c) \Longleftrightarrow f(s)=c$;  & $(C,s) \models \allins \varphi \Longleftrightarrow \forall s' \in S: (C,s') \models \varphi$;\\
  $(C, s) \models \hOp(c_i, c_j) \Longleftrightarrow  c_i \relHierarchy c_j $; & $(C, s) \models \temporalbetter \varphi \Longleftrightarrow \forall s' \in {\leq_T}[s] :  (C,s') \models \varphi $;\\
  $(C, s) \models \bOp(c_i, c_j) \Longleftrightarrow  c_i \relBinding c_j$; & $(C, s) \models \relOp\varphi \Longleftrightarrow \forall s' \in  \relevance(s) :  (C,s') \models \varphi $.\\
  $(C, s) \models\neg\varphi \Longleftrightarrow(C, s) \not\models\varphi $; & 
\end{tabular}

\end{definition}

\noindent A formula $\varphi$
of
$\mathcal{L} (\mathit{Atm})$
is satisfiable
relative
to the class 
$\cmtemp$
if there exists
a pointed classifier model 
$(C,s)$
with $C \in \cmtemp $
such that $(C,s) \models \varphi$. Validity is defined as usual.

By using our unary temporal modality $[\leq_T]$ and name formulas, we can define the binary modalities of future and past --- ``strictly later'' and ``strictly earlier''. The use of names allows us to capture irreflexivity in the temporal dimension. Specifically, we define $\exfuture_{n}\varphi \defin n\wedge \langle \leq_{T}\rangle ( \neg \langle\leq_{T} \rangle   n  \wedge \varphi) $ and $\expast_{n,m}\varphi \defin n \wedge \someins( \varphi \wedge \exfuture_{m}n)$. The operators $\exfuture_{n}$ and $\expast_{n,m}$ act as existential future and past operators --- indicating the existence of a strictly later or strictly earlier point --- as shown in the Proposition below.


\begin{proposition}
I) $(C,s)\models \exfuture_{n}\varphi$ iff $(C,s)\models n$ and $\exists s' \in S$ s.t. $\;s  \stricttemporalorder s'$ and $(C,s')\models  \varphi$.\; II)  $(C,s)\models \expast_{n,m} \varphi$ iff $(C,s)\models n$ and  $\exists s' \in S$ s.t. $(C,s')\models m \wedge \varphi$ and $s'  \stricttemporalorder s$.
\end{proposition}
\vspace{-0.2em}
Further, we can define two new modalities of hierarchy, which are useful to define binding constraints in Section~\ref{sub:Exceptions}: $\Lower \varphi \defin \bigvee_{c,c' \in {\sf Courts}} (c\wedge \hOp(c,c') \wedge \someins (c'\wedge\varphi ))$;
$\Higher \varphi \defin \bigvee_{c,c'\in {\sf Courts}} (c\wedge \hOp(c',c) \wedge \someins (c'\wedge\varphi ))$;
$\Samecourt \varphi \defin\bigvee_{c \in {\sf Courts}} ( c\wedge \someins (c\wedge\varphi ))$. $\Lower \varphi$ reads  ``a lower hierachical state is a $\varphi$-state''. $\Higher \varphi$ reads  ``a higher hierachical state is a $\varphi$-state''.  $\Samecourt \varphi$ reads  ``a state with same court is a $\varphi$-state''. 
\vspace{-1.5em}


\begin{proposition}\label{prop:logicHierarchy}
I) $(C,s)\models \Lower \varphi$ iff there is $s'\in S$ s.t. $s' \stricHierarchy s$ and    $(C,s')\models \varphi$;
II) $(C,s)\models \Higher \varphi$ iff there is $s'\in S$ s.t. $s \stricHierarchy s'$ and    $(C,s')\models \varphi$; III) $(C,s)\models \Samecourt \varphi$ iff there is $s'\in S$ s.t. $s\eqHierarchy s'$ and $(C,s')\models \varphi$.
\end{proposition}

\subsection{Axiomatics and Completeness}
\begin{definition}[Axiomatics of $\logic$]\label{axiomatics}
We define $\logic$
to be the extension of 
propositional logic with 
the following
 axioms and rules:
\end{definition}
\vspace{-30pt}

\vspace{-1.5em}

\begin{table}[h]

\begin{multicols}{2}
\footnotesize{
\begin{align}
\;&\allins \varphi \wedge \allins (\varphi \rightarrow \psi)
\rightarrow \allins \psi
 \tagLabel{K$_{\allins}$}{ax:Kbox}\\
& \allins \varphi \rightarrow  \varphi
 \tagLabel{T$_{\allins}$}{ax:Tbox}\\ 
& \allins \varphi \rightarrow  \allins\allins \varphi
 \tagLabel{4$_{\allins}$}{ax:4box}\\  
& \quad\quad\quad \varphi \rightarrow  \allins \someins \varphi
\tagLabel{B$_{\allins}$}{ax:Bbox}\\
& \bigvee_{o \in \mathit{Val}}\takevalue{o}
 \tagLabel{AtLeastValue}{ax:Leastx}\\
& \takevalue{o}\to \neg \takevalue{o'}\text{ if }o \neq o'
 \tagLabel{AtMostValue}{ax:Mosttx}\\
& \big( \temporalbetter \varphi \wedge \temporalbetter (\varphi \rightarrow \psi) \big)
\rightarrow\temporalbetter\psi
 \tagLabel{K$_{\temporalbetter}$}{ax:Kbett}\\
& \temporalbetter\varphi \rightarrow  \varphi
 \tagLabel{T$_{\temporalbetter}$}{ax:Tbett}\\
& \temporalbetter\varphi \rightarrow  \temporalbetter\temporalbetter \varphi
 \tagLabel{4$_{\temporalbetter}$}{ax:4bett}\\
& @_{n} \big(\temporalbetter \varphi \rightarrow @_{m} \varphi \big) \vee @_{m} \big(\temporalbetter \varphi \rightarrow @_{n} \varphi \big) 
 \tagLabel{Total$_{\temporalbetter}$}{ax:TotalBett}\\
& \allins\varphi \rightarrow  \temporalbetter \varphi
 \tagLabel{MIX$_{ \allins, \temporalbetter}$}{ax:Mboxbett}\\
 & \bigvee_{c_{i}\in \Courts} c_{i}
 \tagLabel{AtLeastCourt}{ax:LeastC} \\
& \court_{i} \to \neg \court_{j}\text{ if }\court_{i} \neq \court_{j}, \court_{i} , \court_{j}\in \Courts
 \tagLabel{AtMostCourt}{ax:MosttC}
\end{align}

\columnbreak

\begin{align}
& \neg \hOp(\court_{i},\court_{i}) 
 \tagLabel{IrrHierarchy}{ax:IrrHier}\\
& \hOp(\court_{i},\court_{j}) \wedge \hOp(\court_{j},\court_{k}) \to \hOp(\court_{i},\court_{k})
 \tagLabel{TrHierarchy}{ax:TrHier}\\
& \hOp(\court_{i},\court_{j})\to \allins\hOp(\court_{i},\court_{j})
 \tagLabel{GlobHier}{ax:GlHier}\\
& \bOp(\court_{i},\court_{j})\to \allins\bOp(\court_{i},\court_{j})
 \tagLabel{GlobBind}{ax:GlBind}\\
& \big( \rOp \varphi \wedge \rOp (\varphi \rightarrow \psi) \big)
\rightarrow \rOp \psi
 \tagLabel{K$_{\rOp}$}{ax:KR}\\
& \allins\varphi \rightarrow  \rOp \varphi
 \tagLabel{MIX$_{ \allins, \rOp}$}{ax:MboxR}\\
& \frac{n\rightarrow \varphi }{ \varphi } \;\text{ where } n \;\text{does not occur in } \varphi
 \tagLabel{NAME}{rule:NAME}\\
& \someins\big( n \wedge \varphi \big) \rightarrow @_n \varphi 
 \tagLabel{nam1}{ax:Name1}\\
& \neg \takevalue{?} \leftrightarrow \bigvee_{n\in \Names_{d}} n
 \tagLabel{nam2}{ax:Name3} \\
& \frac{\phi \to \psi,  \hspace{0.25cm} \phi}{\psi}
\tagLabel{MP}{rule:MP}\\
& \frac{\varphi }{\allins \varphi }
 \tagLabel{Nec$_{\allins}$}{rule:Necbox}
\end{align}}
\end{multicols}
\end{table}
\vspace{-2.5em}




$\allins$ is an S5 operator. $\temporalbetter$ is a S4 operator. \ref{ax:TotalBett} reflects totality of temporal relation. $\relOp$ is a normal operator. \ref{ax:Leastx},  \ref{ax:Mosttx} syntactically represent the decision function  \cite{LiuLoriniJLC}. \ref{ax:TrHier}, \ref{ax:IrrHier}  capture the hierarchical structure. \ref{ax:GlHier} and \ref{ax:GlBind} express the universality of the hierarchical and binding relations.   \ref{ax:LeastC}, \ref{ax:MosttC} ensure the existence and uniqueness of courts. \ref{ax:Name1}, adapted from \cite{gorankoNames}, guarantees that each name uniquely identifies a single state; \ref{ax:Name3} ensures that all and only decided states are associated to names in $\Names_d$, and so the set of decided states is finite;  
  \ref{rule:NAME}, from \cite{Blackburn_Rijke_Venema_2001}, guarantees the construction of canonical models with named states.


We will focus on $\extlogic$ the extension of $\logic$ given by axioms:\vspace{-10pt}
\begin{center}
\begin{tabular}{p{8.2cm}p{1.2cm}p{3cm}p{1.5cm}}
   $\bigwedge_{c_i \neq c_j} \big((\hOp(c_i,c_j) \rightarrow   \bOp(c_i,c_j))\wedge (\neg \hOp(c_i,c_j) \rightarrow  \neg \bOp(c_i,c_j)\big)$  & ({\bf SL}) \label{ax:SL} & $\allins (\takevalue{?}\rightarrow \temporalbetter \takevalue{?})$ & ({\bf OL}) \label{ax:ol}
\end{tabular}
\end{center}



Axioms {\bf SL} and {\bf OL} corresponds respectively to    \ref{eq:SD} and \ref{eq:T}. 

\begin{proposition}
$\logic$  is sound and complete with respect  to $\cmtemp$. $\extlogic$  is sound and complete with respect to $\scmtemp$. 
\end{proposition}

\section{Precedents, Binding Precedents and Exceptions}


A precedent for a case under consideration is a relevant case that was previously decided. However, not all precedents are \emph{potentially binding}. Potentially binding precedents are those  established by a court that holds binding authority over the court  deciding the case at hand.
For example, a precedent set by the UK Supreme Court is potentially binding on the High Court --- but the reverse does not hold.
We use the term \emph{potentially binding precedent} because such precedents are subject to exceptions: 
they may lose their binding force in certain circumstances.
Indeed, a case may be \emph{overruled} by a court with the power to do so; or it may have been decided in disregard of a binding precedent without the authority to do so, and thus be considered \emph{per incuriam} \cite{LexisNexis}. Accurately modeling the notion of exceptions will require particular care. 

\vspace{-0.3cm}

\subsection{Potentially Binding precedents}\label{sub:potBind}


A precedent for a case at hand is a relevant case that was decided earlier. Specifically, $s'$ is a supporting precedent --- or simply a precedent --- for $s$ if $s'$ is relevant for $s$ and was decided before $s$. For $s,s',s''\in S$, we denote their courts $c,c',c''$,  (e.g. $c\in s\cap \Courts$).

\begin{definition}[Precedent]\label{def:support}
 Let $s,s'\in S$ and $\outcome\in \{0,1\}$. $s'$ is a (supporting) precedent for $s$ in the direction of $\outcome$, noted $\Pi(s',s,\outcome)$, iff $f(s')=\outcome$, $s' \in \relevance( s )$, and $s'<_{T}s$.
\end{definition}

Supporting precedents are the relevant past cases. In this context, we use a hybrid style to identify which states qualify as precedents, represented by using the intersection of both relevance and temporal modalities, as illustrated below.







\begin{definition}
The notion that for the $n$-named state a supporting state is $m$-named and a $\varphi$-state, denoted as $\mathsf{Supporting}_{n,m}\varphi$, is defined as $\expast_{n,m}\varphi \wedge \relexOp(\varphi\wedge m)$. 
\end{definition}





\begin{proposition}\label{prop:Supporting}
Let 
$\outcome \in \{0,1\}$. So, $(C,s)\models \mathsf{Supporting}_{n,m}( \takevalue{\outcome} \wedge \varphi)$ iff  $(C,s)\models n$ and exists $s'\in S$,  s.t.  $ \;\Pi(s',s,\outcome)$ and  $(C,s')\models m \wedge \varphi$. 
\end{proposition}

\begin{example}
$(C_{ex}, s^{*})\models \mathsf{Supporting}_{n*,n_{1}}\takevalue{1}$: $s_{1}$ named $n_{1}$ is decided as $1$ and is relevant and before $s^{*}$, named $n^{*}$.  Similarly, $(C_{ex},s^{*})\models\mathsf{Supporting}_{n*,n_{2}}\takevalue{1}\wedge\mathsf{Supporting}_{n*,n_{3}}\takevalue{0}\wedge \mathsf{Supporting}_{n*,n_{4}}\takevalue{1}\wedge \mathsf{Supporting}_{n*,n_{5}}\takevalue{0}\wedge \mathsf{Supporting}_{n*,n_{6}}\takevalue{1}$.  
\end{example}

Not all supporting precedents are potentially binding, i.e. not all of them may force the decision in current case. The potentially binding precedents for a state $s$, decided by court $\court$, are those precedents  $s'$ issued by a court $\court'$ that holds binding authority over $\court$.  



\begin{definition}[Pot. bind.]\label{def:bindingprec}
Let $s, s'\in S$ and $\outcome\in \{0,1\}$.
$s'$ is potentially a binding precedent for $s$ for a decision as $\outcome$, denoted as $\beta (s',s,\outcome)$, iff $\Pi(s',s,\outcome) \text{ and } \court' \relBinding \court$. We simply write $\beta (s',s)$ iff there is $\outcome\in \{0,1\}$ s.t. $\beta (s',s,\outcome)$.  We define $\beta_{s}=\{s'\mid \beta (s',s) \}$. 
\end{definition}

\begin{definition}
For the $n$-named state a potentially binding state is a $m$-named $\varphi$-state, noted as $\mathsf{PBinding}_{n,m}\varphi$, is defined as $\bigvee_{c_i, c_j\in \Courts} \Big(\bOp(c_i, c_j) \wedge  (\mathsf{Supporting}_{n,m}\big(c_i \wedge \varphi) \Big)$. 



\end{definition}

\begin{proposition}\label{prop:Binding}
Let $s\in S$, $\outcome \in \{0,1\}$.  $(C,s)\models \pbinding_{m,n}( \takevalue{\outcome}\wedge \varphi)$ iff  $(C,s)\models n$ and exists $s'\in S$,  s.t $\;\beta(s',s,\outcome)$ and $(C,s')\models m\wedge  \varphi$.
\end{proposition}

\begin{example}\label{ex:pbinding}
$(C_{ex}, s^{*})\models \pbinding_{n*,n_{1}}\takevalue{1}$ means: $s_{1}$ is supporting for $s^{*}$ and decided by court $c_{0}$, which is binding for court $c_{2}$ in $s^{*}$. Same can be verified for $s_{2}$, $s_{4}$ (with $1$) and $s_{3}$ (with $0$). However, 
$(C_{ex}, s^{*})\not\models \pbinding_{n*,n_{5}}\takevalue{0}$. Although $s_{5}$ is supporting for $s^{*}$, $s_{5}$ is decided by $c_{4}$ which is not binding for $s^{*}$ (same can be verified for $s_{6}$).
\end{example}



\subsection{Exceptions: Incuriam and Overruling}\label{sub:Exceptions}

\vspace{-0.7cm}

    \begin{SCfigure}[0.7][htb]
    {\begin{tikzpicture}[scale=0.7, transform shape,->, node distance=2.6cm, thick]

\node (Sstar) {$s^*$};
\node (S1) [right of=Sstar] {$s_1$};
\node (S2) [right of=S1] {$s_2$};
\node (S3) [right of=S2] {$s_3$};

\draw (S1) -- (Sstar) node[midway, above] {p-binding};
\draw (S2) -- (S1) node[midway, above] {overrules};
\draw (S2) -- (S3) node[midway, above] {per incuriam};

\end{tikzpicture}
}
\caption*{Figure D}
\label{plates1}
    \end{SCfigure}
\vspace{-1.3cm}

    \begin{SCfigure}[0.7][htb]
    {\begin{tikzpicture}[scale=0.7, transform shape,->, node distance=2.6cm, thick]

\node (Sstar) {$s^*$};
\node (S1) [right of=Sstar] {$s_1$};
\node (S2) [right of=S1] {$s_2$};
\node (S3) [right of=S2] {$s_3$};

\draw (S1) -- (Sstar) node[midway, above] {p-binding};
\draw (S1) -- (S2) node[midway, above] {per incuriam};
\draw (S3) -- (S2) node[midway, above] {overrules};

\end{tikzpicture}
}
\caption*{Figure E}
\label{plates1}
    \end{SCfigure}

\vspace{-0.7cm}


Potentially binding precedents are subject to two  exceptions: they may be overruled or decided \emph{per incuriam}. Intuitively, a case $s$ is overruled when a later, relevant case is decided differently by a court that holds overruling power with respect to the court in $s$. Conversely, a case $s$ is decided \emph{per incuriam} if it goes against a binding precedent, without having authority to do so. To formally define these exceptions we begin by considering these aspects:
\begin{varitemize}
\item   \emph{Per incuriam} and overruled cases lose their bindingness on subsequent cases, i.e. we can remove them from  binding precedents.
\item The model does not specify which cases are \emph{per incuriam} or overruled. Rather, as shown later in this section, this information is computed through a recursive process involving  the following interactions between \emph{per incuriam} and overruling.
\item  Consider Figure D. Suppose we have a case $s^{*}$ to be decided, for which $s_{1}$ is a potentially binding precedent. Now assume that $s_{2}$ --- for which $s_{1}$ was relevant --- is decided later and differently by a court with overruling authority. Intuitively, this suggests that $s_{1}$ has been overruled and should no longer be binding  for $s^{*}$. However, suppose we then discover that $s_{2}$ itself contradicts a binding precedent $s_{3}$, without having the authority to do so --- that is, $s_{2}$ was decided \emph{per incuriam}. In that case, $s_{2}$ was not binding on $s_{1}$, and we can reinstate $s_{1}$ as a binding precedent for $s^{*}$. 
Hence, a case loses its overruling power when it is \emph{per incuriam.} 
\item Consider Figure E. Suppose we have a case $s^{*}$ to be decided, and a potentially binding precedent $s_{1}$. Now assume that $s_{1}$  went against $s_{2}$ without  the authority to do so. Intuitively, this would suggest that $s_{1}$  \emph{per incuriam} and should be excluded from the binding precedents for $s^{*}$. However, suppose we later discover that $s_{2}$ had already been overruled by $s_{3}$ before $s_{1}$ was decided. In that case, $s_{2}$ was no longer binding at the time of $s_{1}$'s decision, and $s_{1}$ can be ``reinstated'' as a valid binding precedent for $s^{*}$. 
A case is not \emph{per incuriam} if it goes against  an overruled case.
\end{varitemize}

\noindent
To capture such interactions, we proceed as follows. We first identify potential overruling cases --- those that can overrule others unless deemed \emph{per incuriam}. We then formally define \emph{per incuriam}. Finally, we define overruled cases.

Not all courts have the authority to overrule prior decisions. 
Within common law systems typically higher courts can overrule lower courts decision. Viceversa, lower court cannot  overrule a higher court. 
More attention requires self-overruling: intuitively, if a court can overrule its own previous decisions, then it is not bound by its own previous decisions.  
So we assume that a court $c'$ may overrule a decision made by court $c$ when deciding  $s'$, if: (1) $c'$ is hierarchically higher to $c$; or (2) $c' = c$ and not-$c\relBinding c$. 



\begin{definition}[Overruling Power]\label{def:overrulingpower}
Court $c'$ has the power to overrule (a decision  by) court $c$, denoted  $O (c', c )$, iff $\relHierarchy (c',c) \text{ or } (c'= c \text{ and not-}c\relBinding c$).
\end{definition}

\begin{definition} 
That a decision made by $c_i$ can be overruled in current state, denoted as $\mathsf{PwOver}(c_i)$, is defined as $\bigvee_{  c_j\neq c_i } \big(c_j \wedge \hOp (c_j, c_i) \big) \vee \big(c_i \wedge \neg\bOp(c_i,c_i))\big)$.
\end{definition}

\begin{proposition}\label{prop:PwOver}
Let $c \in s$. So, $(C,s)\models\mathsf{PwOver}(c_i)$ iff $O(c, c_i)$. 
\end{proposition}



 We now define potential overruling cases. Intuitively, $s'$ may overrule a precedent $s$, when: 1) $s'$ is decided by $c'$ in the opposite direction wrt $s$ and 2) $c'$ has overruling power over $c$ when deciding $s'$ (i.e. $O (c', c )$). 
\begin{definition}[Pot. Over.]\label{def: overruledcase} 
The case $s'$ potentially overrules  $s$, denoted as $O(s',s)$, iff $\Pi(s,s',\outcome)$, $f(s')= \overline{\outcome}$, and $O(c',c)$.  We write  $\omega_{s}=\{s'\mid O (s',s) \}$ for the set of potentially overruling states wrt. $s$. Given a case $\tilde{s}\in S$, we write $Overrule_{T}(s,\tilde{s})=\{s'\mid  O(s',s), \; s'<_{T}\tilde{s}\}$ to represent the set of states potentially overruling $s$ and that were assessed before $\tilde{s}$ was decided. 
\end{definition}

\begin{remark}\label{remark:selfbind}
It can be verified that: If a case $s$ is potentially bounded by case $s'$, then $s$ cannot potentially overule $s'$. Namely, $s'\in \beta_{s}  \Rightarrow s\not \in \omega_{s'}$.
\end{remark}


\begin{definition}
 The notion that for the $n$-named state a potentially overruling state is $m$-named and a $\varphi$-state, denoted as $\poverruling_{n,m}\varphi $, is defined as  $ \bigvee_{\outcome \in \{0,1\},\court} \Big( n \wedge c \wedge \someins\big(\varphi \wedge \takevalue{\overline{o}}\wedge \mathsf{PwOver}(c)\wedge 
\mathsf{Supporting}_{m,n}\takevalue{o}\big)\Big)$.
\end{definition}

\begin{proposition}\label{prop:poverruling}
$(C,s)\models \poverruling_{n,m}\varphi$ iff  $(C,s)\models n$ and exists $s'\in S$ s.t.  $O(s',s)$ and $(C,s')\models m\wedge \varphi$.
\end{proposition}

\begin{example}\label{ex:potOVer}
$(C_{ex},s_{3})\models\mathsf{PwOver}(c_0)\wedge \takevalue{0}$ means: $c_{0}\in s_{3}$,  $c_0$ is not self-bound, and $(C_{ex},s_3)\models \takevalue{0}$. It can be verified that $(C_{ex},s_{3})\models \mathsf{Supporting}_{n_3, n_1}(\takevalue{1})$. Hence   $(C_{ex}, s_{1})\models \poverruling_{n_{1},n_{3}}\takevalue{0}$ --- $s_{3}$ potentially overruled  $s_{1}$. 
\end{example}



\noindent Intuitively, a \emph{per incuriam} case is a case that went \emph{against} a binding precedent. To model this, we define   the notions of going against/according a binding precedent. 

\begin{definition}\label{def:against}
The case $s$ went against $s'$, denoted as $Against(s,s')$, iff $f(s)=\outcome\in\{0,1\}$ and $\beta(s',s,\overline{\outcome})$. The case $s$ was decided according $s'$, denoted as $According(s,s')$, iff $f(s)=\outcome\in\{0,1\} $ and $ \beta(s',s,\outcome)$.
\end{definition}


\begin{definition}\label{prop:AgainstSem}
The notion that the n-named state went against a m-named and potentially binding $\varphi$-state, denoted as $\mathsf{Against}_{n,m}\varphi$, is defined as $\bigvee_{\outcome\in\{0,1\}} \Big( \takevalue{\outcome}\wedge  \pbinding_{n,m}(\varphi\wedge \takevalue{\overline{\outcome}})\Big)$.
The n-named state is decided according to a m-named and potentially binding $\varphi$-state, denoted as $\mathsf{According}_{n,m}\varphi$, is defined  $\bigvee_{\outcome\in\{0,1\}} \Big( \takevalue{\outcome}\wedge  \pbinding_{n,m} ( \varphi\wedge \takevalue{\outcome})\Big).$
 \end{definition}

%

\begin{proposition}\label{prop:Against}
I) $(C,s)\models \mathsf{Against}_{n,m}\varphi$  iff $(C,s)\models n$ and $\exists s'$ s.t. $Against(s,s')$ and $(C,s')\models m \wedge \varphi$; II) $(C,s)\models \mathsf{According}_{n,m}\varphi$ iff $(C,s)\models n$ and $\exists s'$ s.t. $According(s,s')$ and $(C,s')\models m \wedge \varphi$.
\end{proposition}

Consider a state $s$ that went against a binding precedent $s'$, i.e. $Against(s,s')$.  By Remark \ref{remark:selfbind},  $s$ had no overruling power wrt $s'$.  This suggests that $s$ was decided \emph{per incuriam}. However, to properly establish this, four interrelated aspects must be considered.

\begin{varitemize}
\item[1)] Not all courts can   disregard a previous decision     \emph{per incuriam}  \cite{LexisNexis}.
We  assume that a lower court  cannot disregard a  \emph{per incuriam}   precedent by a higher court, to which it  remains bound.~\footnote{In Cassell v Broome (UK), the Court of Appeal was held bound by \emph{per incuriam} House of Lords decision \cite{interpreting}. Canada  follows  same principle \cite{canada}.} But, a court may disregard its own \emph{per incuriam} precedents.~\footnote{Clearly stated for UK Court of App. in \emph{Young v. Bristol Aeroplane Co.}.}  
\item[2)] As mentioned, we  infer from the model which cases are decided \emph{per incuriam}, such  information is not  given. Returning to our question: if $Against(s, s')$, does it mean s was decided \emph{per incuriam}? It depends.
If $s'$ was decided by a higher court, then $s$ is necessarily \emph{per incuriam}, since it lacks the authority to disregard $s'$. If $s$ and $s'$ were decided by the same court, then the court --- when deciding $s$ --- may have found that $s'$ went   against a binding $\overline{s}$. But to confirm that $s'$ was \emph{per incuriam} --- so not binding for $s$ --- we have to check whether $\overline{s}$ was \emph{per incuriam}, etc. Thus, determining if $s$ is \emph{per incuriam} requires tracing its full chain of binding precedents.
\item [3)] Using the same reasoning done for Figure E, we state that $s$ is not \emph{per incuriam} if $s'$ was overruled by another case $\tilde{s}$. The reason is that, if $s'$ was overruled, then it no longer held binding authority over $s$. However, we must also require that $\tilde{s}$ was decided prior to $s$ --- otherwise, $s'$ had not yet been overruled at the time $s$ was assessed and thus $s'$ still  bound  $s$. Moreover, $\tilde{s}$ must itself not be \emph{per incuriam}; otherwise, it would not represent a legitimate overruling. 
\item[4)] For a given case $s$, there may exist conflicting binding precedents: $s'$ and $s''$ within $\beta_s$, such that $Against(s, s')$ and $According(s, s'')$ hold. In this scenario, any decision made in $s$ would have contradicted  one potentially binding precedent. We argue that $s$ can be considered \emph{per incuriam} if $s''$ fails to satisfy the previously established criteria --- namely, if $s''$ is itself \emph{per incuriam} and by a court not higher than $s$, or if it was overruled (prior to $s$) by a non-\emph{per incuriam} precedent. If $s''$ fits neither category, there remains another case in which $s$ may still be regarded as \emph{per incuriam}: following a principle for resolving conflicts of precedents (to be detailed later), $s$ should have followed the later precedent from the higher court. Therefore, we examine the temporal and hierarchical relationship between $s'$ and $s''$. $s$ qualifies as \emph{per incuriam} if: $s''$ was decided by a lower court than $s'$, or by the same court but strictly before.

\end{varitemize}

%
As observed, determining whether a case was decided \emph{per incuriam} is a chain-like evaluation process of binding precedents. Here, a graph $G_{s}=(V_{s},E_{s})$ constructed recursively is given to compute whether a case $s$ was decided \emph{per incuriam.} At step 0, the graph ($G_{0}=(V_{0}, E_{0})$) has only one node $s$ ($V_{0}=\{s\}$) and no edge ($E_{0}=\emptyset$).
At step 1, we add to the nodes  the  $s'$,  the potentially binding precedents or the  potentially overruling states for $s$ ($V_{1}= \{s\}\cup \{s'\mid s'\in  \beta_{s} \text{or} \; s'\in  \omega_{s}\}$). 
Edges joining  $s$ to each $s'$ are added ($E_{1}=\{(s, s')\mid   s'\in  \beta_{s}\;  \text{or}\; s'\in  \omega_{s}\}$). We repeat the procedure  recursively for each node.



\begin{definition}(s-Graph)\label{def:Sgraph}
A structure $G_{s}= (V_{s}, E_{s})= \bigcup_{n\geq 0} G_{n}$ is called a $s$-graph, when $f(s) \neq ?$ and every $G_{n}$ satisfies: 
\vspace{-0.3cm}
\begin{itemize}
\item $G_{0}= (V_{0}, E_{0})$, where $V_{0}=\{s\}$ and $E_{0}= \emptyset$;
\item $G_{n+1}= (V_{n+1}, E_{n+1})$, where 
\begin{itemize}
\item $V_{n+1}=V_{n}\cup \{s'\mid s'\in  \beta_{s''} \text{ or }  s'\in  \omega_{s''}, \text{ with } s'' \in  V_{n}  \}$ and 
\item $E_{n+1}=E_{n}\cup \{(s'',s')\mid  s'\in  \beta_{s''} \text{ or }  s'\in  \omega_{s''}, \text{ with } s'' \in  V_{n}\}.$
\end{itemize}
\end{itemize}
    
\end{definition}
\vspace{-0.3cm}
\noindent
$G_s$ is a finite directed acyclic graph. We thus define the \emph{height} of a state $s$, denoted $\text{height}(s)$, as the number of edges in the longest path from $s$ to a sink node in $G_s$.~\footnote{A sink node is a node with no outgoing edges.}



By recursively exploring graph $G_{s}$, we can compute whether $s$ is \emph{per incuriam} (i.e. $Incuriam(s)$). We claim that $s$ is  \emph{per incuriam}  if: 
 \begin{varitemize}
 \item  there is $s'$ adjacent node to $s$ (i.e. $(s,s')\in E_{s}$), such that: a) $s$ is  against $s'$ (i.e. $Against(s,s')$);   b) if $s'$ was decided  \emph{per incuriam}  then it is hiearchically superior to $s$ (i.e. $Incuriam (s') \Rightarrow  s\stricHierarchy s'$); and c) every $\tilde{s}$ potentially overruling $s'$ (before $s$) is itself  \emph{per incuriam} (i.e. $\forall \tilde{s}\in Overrule_{T}(s',s):\;Incuriam(\tilde{s
})$).
 \item for all $s''$, adjacent node to $s$ (i.e. $(s,s'')\in E_{s}$) s.t. $According(s,s'')$  we have:  a) $s''$ was decided  \emph{per incuriam} by a court that is not higher than the court of $s$ (i.e. $Incuriam (s'') \text{ and }  s \not\stricHierarchy s''$) ); or b) there is  $\tilde{s}$ potentially overruling  $s'$ (before $s$) and not \emph{per incuriam} (i.e. $\exists \tilde{s}\in Overrule_{T}(s'',s)$ s.t. not $Incuriam(\tilde{s})$); or  c) $s''$ was decided by a lower court of $s'$ or $s''$ was decided by the same court of $s'$ but strictly before (i.e. ($s''<_{T} s'$ and $s' \eqHierarchy s''$ )  or $s''\stricHierarchy s'$).
 \end{varitemize}

\begin{definition}(Incuriam)
    Let $G_{s}=(V_{s}, E_{s})$ be a $s$-graph. The state $s$ was decided \emph{per incuriam}, denoted as $Incuriam (s)$, iff \vspace{-5pt}
\begin{itemize}
\item $\exists  s'  \in S$ s.t. $(s,s')\in E_{s}$, $Against(s,s')$, ($Incuriam (s') \Rightarrow  s\stricHierarchy s'$), and $\forall \tilde{s}\in Overrule_{T}(s',s):\;Incuriam(\tilde{s
})$.
\item $\forall  s'' \in S$ if $(s,s'')\in E_{s}$ and $According(s,s'')$ then either [$Incuriam (s'')$ and $s \not\stricHierarchy s''$], or [$\exists \tilde{s}\in Overrule_{T}(s'',s) : not \;Incuriam(\tilde{s})$], or [($s''<_{T} s'$ and $s' \eqHierarchy s''$) or $s''\stricHierarchy s'$].
\end{itemize}
\end{definition}


We now use our modal language to represent the notion of \emph{per incuriam}. As previously shown, \emph{per incuriam} involves temporal and jurisdictional hierarchies, as well as potential overruling. Given a finite set $\Names_{d}$ of decided names, we define the following useful formulas to express \emph{per incuriam}: $\exfuture\varphi \defin \bigvee_{n\in \Names_{d}}\big( n\wedge  \exfuture_{n}\varphi\big)$;  $\mathsf{Against}\varphi \defin \bigvee_{n, m \in \Names_{d}}\big( n \wedge \mathsf{Against}_{n,m}\varphi\big)$; $\mathsf{POverruling}\varphi \defin \bigvee_{n, m\in \Names_{d}}\big( n\wedge \mathsf{POverruling}_{n,m}\varphi\big)$;  $\forallaccording\varphi \defin \bigvee_{n\in \Names_{d}}\big( n \wedge \bigwedge_{m\in \Names_{d}}\neg \mathsf{According}_{n, m}\neg \varphi\big)$;  $\forallpoverruling\varphi \defin \bigvee_{n\in \Names_{d}}\big( n\wedge \bigwedge_{m\in \Names_{d}}\neg \mathsf{POverruling}_{n, m}\neg \varphi\big)$. They will be used to define the expressions that characterize the recursive nature of computing whether a case was decided \emph{per incuriam}.

\begin{definition}
We define formula $\iota^{n}$, recursively on $n\in \mathbb{N}$: 
{\footnotesize
\begin{alignat*}{3}
\iota^{1}\defin& 
\bigvee_{m\in \Names_{d}} &&\Big(\mathsf{Against}(m) \wedge \forallaccording\big((\exfuture m \wedge \Samecourt (m))\vee \Higher(m)\big)\Big),\\
\iota^{n+1}\defin &\bigvee_{n,m\in \Names_{d}} &&\Big(n\wedge  \mathsf{Against} \big(m\wedge(\neg \iota^{n}\vee \Lower(n)) \wedge \forallpoverruling(\exfuture n \rightarrow \iota^{n} )\big)
\wedge \\
&\;&&\forallaccording\big((\iota^{n} \wedge \neg \Lower(m)) 
\vee \mathsf{POverruling}(\exfuture n \wedge \neg \iota^{n} ) \vee((\exfuture m \wedge\Samecourt (m))\vee \Higher(m))  \big)\Big).
\end{alignat*}}
\end{definition}
\vspace{-0.2cm}
Let $h= height(s)$. It can be proven that if a state satisfies a $\iota^{k}$ for some $k\geq h=height(s)$, then it satisfies $\iota^{j}$ for all $j\geq h=height(s)$:

\begin{equation} \tag{$\star$}\label{cond:sol}
 \text {If } \exists k\geq h: (C,s)\models \iota^{k} \text { then } \forall j\geq h: (C,s)\models \iota^{j}.
\end{equation}

We can finally define the notion of \emph{per incuriam} in the language. 


\begin{definition}
$\mathsf{Incuriam}\defin \bigvee_{k \leq |\Names_{d}| } \bigwedge_{ k\leq j\leq |\Names_{d}| } \iota^{j}$. 
\end{definition}



We aim to show that  semantic and syntactic definitions of  
\emph{per incuriam} are equivalent, i.e. $Incuriam(s)$ iff exists $k\leq |\Names_{d}|$ s.t. $s$ satisfies  $\iota^{j}$, for all  $j$ with $|\Names_{d}|\geq j\geq k$. We claim  $k$ is actually \emph{at most} the height  $s$ ($h=height(s)$)\footnote{$h\leq |\Names_{d}|$:   states of $G_{s}$ are in $S_{d}$ and we can  prove that $|S_{d}|\leq |\Names_{d}|$.}. The formal proof is by induction; here, we provide hints focusing on 
   $(C,s)\models \mathsf{Incuriam}\Rightarrow Incuriam(s)$. 
\begin{varitemize}
\item  Assume $(C,s)\models \mathsf{Incuriam}$. Then $(C,s)\models\iota^{h}$, exactly at $h=height(s)$.  Indeed, by definition of $ \mathsf{Incuriam}$,  there is $k\leq |\Names_{d}|$  s.t. $(C,s)\models \iota^{j}$, with  $|\Names_{d}|\geq j\geq k$. If $k\leq h$, then trivially $(C,s)\models \iota^{h}$. If $k> h=height(s)$,   (\ref{cond:sol}) guarantees that $(C,s)\models \iota^{h}$.
 \item So, if we want to prove  $(C,s)\models \mathsf{Incuriam}\Rightarrow Incuriam(s)$ we just need to prove that $(C,s)\models \iota^{h} \Rightarrow  Incuriam(s)$, for $h=height(s)$.  
 \item We give an intuition that previous point holds for $h=1$. Suppose $(C,s)\models \iota^{1} $, from this we can verify 1) there exists a state $s'$ s.t. $Against(s,s')$ and  2) for all $s’'$ s.t. $According(s,s'')$: ($s\stricHierarchy s'$ or ($s'\eqHierarchy s''$ and $s'\stricttemporalorder s''$)). Is this enough to say $Incuriam(s)$? Yes --- provided that $Incuriam(s')$  and $s'$ has not been overruled. This is the case. Indeed, since $height(s)=1$ then $height(s')=0$ --- from $Against(s',s)$, we know $s'\in \beta_{s}$, so $s'$ is a successor of $s$ in the graph. Since $height(s')=0$, $s'$ has no successors in the graph. This means  $s'$ admits neither binding precedents nor potentially overruling cases.  Hence,  not $Incuriam(s')$ and $s'$ is not overruled.
 \end{varitemize}

\begin{proposition}\label{prop:IncuriamSemLog}
Let $s\in S$. $(C,s)\models \mathsf{Incuriam} \Leftrightarrow Incuriam(s)$.
\end{proposition}

We have modeled \emph{per incuriam}. We now turn to another exception: overruled states --- those for which there exists a potentially overruling state that is not \emph{per incuriam}.

\begin{definition}
$Overruled(s)$ iff there is $s'\in \omega_{s}$ s.t. not $Incuriam(s')$. 

$\mathsf{Overruled}\defin \mathsf{POverruling}(\neg \mathsf{Incuriam})$. 
\end{definition}

\begin{proposition}\label{prop:overruledSemLog}
    $(C,s)\models \mathsf{Overruled} $ iff $Overruled(s)$.
\end{proposition}

Finally, we arrive at the notion of binding precedents \emph{without exception}: potentially binding precedents that are neither \emph{per incuriam} (by same court) nor overruled.

\begin{definition}[Binding]
$s'$ is a  binding precedent (without exception) for $s$ iff $s'\in \beta_{s}$ and  not ($Incuriam(s)$ and $s\eqHierarchy s'$) and not $Overruled(s')$.  $\overline{\beta}_{s}$ is the set of binding precedents  without exception for $s$.  
    \end{definition}
    
    \begin{definition}
The notion that  for the n-named state a binding state without exception is m-named and a $\varphi$-state, denoted $\mathsf{Binding}_{n}\varphi$, is defined as $\bigvee_{m\in \Names_{d}}\mathsf{PBinding}_{n,m}\big( \varphi\wedge\neg \mathsf{Overruled} \wedge \neg (\mathsf{Incuriam}\wedge \Samecourt(n))\big)$.
\end{definition}


As shown in the proposition below, our modal language can be used to represent the concept of binding precedents without exceptions.

\begin{proposition}\label{prop:BindingSemLog}
$(C,s)\models  \mathsf{Binding}_{n}$ iff $(C,s)\models n$ and there is $s'\in  \overline{\beta}_{s}$ s.t. $(C,s')\models \varphi$. 
\end{proposition}

\begin{example}\label{ex:noexceptions}
By Ex. \ref{ex:pbinding}, $s_1$, $s_2$, $s_3$, $s_4$ are potentially binding for $s^{*}$. $s_4$ is \emph{per incuriam}, i.e. $(C_{ex}, s_{4})\models  \mathsf{Incuriam} $. Indeed, $(C_{ex},s_{4})\models\iota^{j}$ for all $j\geq height(s_{4})$, where $height(s_{4})=1$ (the $s_{4}$-graph contains only $s_{4}$ and its binding  $s_{3}$ --- that in turn has no binding precedent nor potentially overruling case). By (\ref{cond:sol}), we   need to verify that $(C_{ex}, s_{4})\models \iota^{1}$. This holds since $s_4$ goes against $s_3$ and $s_4$ has no according state. 
It can also be verified that $s_{1}$ is overruled by $s_{3}$. Also, $s_{2}$ and $s_{3}$ are not \emph{per incuriam} or overruled. So, the binding precedents without exception for $s^{*}$ are $s_{2}$ and $s_{3}$, decided resp. as $1$ and $0$:  $(C_{ex},s^{*})\models\mathsf{Binding}_{n*}(\takevalue{1})\wedge \mathsf{Binding}_{n*}(\takevalue{0})$. We need a principle to solve  the conflict.  
\end{example}

\vspace{-0.5cm}

\section{Resolving Precedents Conflict for New Cases}\label{sec:newClass}

When binding precedents conflict, the most recent decision from the highest court should be followed --- we call this  the \emph{Temporal Hierarchical Principle}. It reflects the idea that political, economic, or social changes may influence a court’s stance \cite{interpreting}. U.S. courts, for example, adopt this principle \cite{interpreting,Broughton}. Here, we formalise a decision process for a new case $s^{*}$ (i.e. $f(s^{*})=?$) based on this principle.  First we define the best temporal hierarchical binding precedents --- those more recent and from higher courts.

   \begin{definition}
The set of best temporal hierarchical binding precedents for $s\in S$ is $Best_{TH}(\overline{\beta}_{s}) = \{s'\in\overline{\beta_{s}}\mid \forall  s''\in \overline{\beta}_{s}$: $s'\not\stricHierarchy s''$ and not ($s'\eqHierarchy s''$ and $s'\stricttemporalorder s'')\}$. 
 \end{definition}


 \begin{definition} 
  That for the n-named state, there is  a best temporal hierarchial binding state without exception in which  $\varphi$ holds, denoted as $\mathsf{BestBinding}_{n}\varphi$, is defined as 
  \vspace{-0.18cm}
 \begin{align*}
\footnotesize{\bigvee_{m\in \Names_{d}}} \Big(&\mathsf{Binding}_{n}( \;m \wedge \varphi) \wedge \neg \mathsf{Binding}_{n}\big((\Lower(m))\vee
(\Samecourt (m) \wedge\expast m)\big) \Big).
\end{align*}
\end{definition}

 \vspace{-0.2cm}
The best binding precedents thus can be expressed by our modal formulas, as shown above, capturing those that satisfy \emph{Temporal Hierarchical Principle}.

 \begin{proposition}\label{prop:THbestbinding}
 $(C,s)\models \mathsf{BestBinding}_{n}\varphi$ iff $(C,s)\models n$   and  $\exists s'\in Best_{TH}(\overline{\beta}_{s})$ s.t. $\footnotesize{(C,s')\models\varphi}$.
 \end{proposition}

 

We define a decision making process based on the \emph{Temporal Hierarchical Principle}.

\begin{definition}[Temp. Hier. Principle]
Let $s^{*}$ s.t. $f(s^{*})=?$. The decision making function $f^*: S \to 2^{\{0,1, ?\}}$, based on best temporal hierarchical binding precedents is \vspace{-6pt}
    \begin{equation*}
f^{*}(s) =
        \begin{cases}
\{f(s)\}  &\text{ if } s\neq s^{*},\\ 
\{f(s')\mid s'\in Best_{TH}(\overline{\beta}_{s^{*}}) \} &\text{ otherwise. }
    \end{cases}
    \end{equation*}
    \end{definition} \vspace{-2pt}
\noindent    
The decision process $f^{*}$ assigns to $s^{*}$ the decisions among the best temporal hierarchical binding precedents in $Best_{TH}(\overline{\beta}_{s^{*}})$. Note that there may be two such precedents with conflicting outcomes --- two binding precedents decided differently (as $0$ and as $1$), by higher courts and considered to apply ``simultaneously''. In such a case, the decision process may yield both outcomes, $f^{*}(s^{*})=\{0,1\}$. Conversely, if all the best binding precedents agree on a single outcome $\outcome \in \{0,1\}$, we say that the decision is \emph{forced} to $\outcome$, namely $f^{*}(s^{*})=\{\outcome\}$. Both scenarios can be expressed using our  language. The formula $\mathsf{BestBinding}_{n*}(\takevalue{\outcome}) \wedge  \mathsf{BestBinding}_{n*}(\takevalue{\overline{\outcome}})$ holds for $s^{*}$ in the first scenario. While, in the second scenario $\mathsf{BestBinding}_{n*}(\takevalue{\outcome}) \wedge  \neg \mathsf{BestBinding}_{n*}(\takevalue{\overline{\outcome}})$ holds for $s^{*}$.



\begin{definition}
The notion that for the n-named state the decision is forced to be $\outcome \in \{0,1\}$, denoted $\mathsf{Cl}_{n}(\outcome)$, is defined as $\mathsf{BestBinding}_{n}(\takevalue{\outcome}) \wedge  \neg\mathsf{BestBinding}_{n}(\takevalue{\overline{\outcome}})$.
\end{definition}

\begin{proposition}
$(C,s^{*})\models \mathsf{Cl}_{n}(\outcome) $ iff $(C,s^{*})\models n$ and $f^{*}(s
^{*})=\{\outcome\}$.
\end{proposition}

 \begin{example}
$(C_{ex},s^{*})\models  \mathsf{BestBinding}_{n*}(\takevalue{0} )\wedge \neg  \mathsf{BestBinding}_{n*}(\takevalue{1}) $ :  $s_2$, $s_3$ are binding without exception for $s^{*}$. $s_{3}$ is decided (as $0$), by a higher court than  $s_2$. So, $(C_{ex},s^{*})\models  \mathsf{Cl}_{n*}(0) $.  $s^{*}$ is forced to be $0$.  We have solved the conflict in Example \ref{ex:noexceptions}. 
 \end{example}

\vspace{-0.25cm}

\section{Conclusion and Related Works}

We proposed an extension of BCL logic for classifiers \cite{LiuLoriniJLC}, by introducing new modalities that capture the relevance relationships between cases, their temporal ordering, and the hierarchical structure of the legal system under consideration. We  modeled the notion of precedents, including binding precedents and those that admit exceptions --- such as precedents decided \emph{per incuriam} or  overruled. Finally, we formalised within the language a temporal and hierarchical principle for resolving conflicts of precedents.

The relevance relation  merits further attention. We imposed no  properties on the relevance relation. As shown in~\cite{PrecedentsClash}, the relevance relation derived from Horty result model is neither transitive nor reflexive nor symmetric. Also, leaving relevance unconstrained allows for cases that are both relevant to a new case but not to each other --- a feature important for modeling decisions involving multiple concerns (see \cite{branting,capon21,CanovottoHierarchies}).

Also, the notion of binding is  “binary": a court  issues  binding decisions or not. But, sometimes,  as suggested in  \cite{deonticAuth}, a binding relation  between two agents of an institution may be context-dependent, as it happens for the UK supreme court (see footnote 1). In the future, we plan to consider a more sophisticated binding relation. 

Other than \cite{liu2022modelling}, no modal logics for legal CBR are known to us.
Nonetheless, two prior works explored the role of time and hierarchy in legal CBR, from a semantic perspective. \cite{Broughton} combines two models of constraint, based on the assumption that vertical constraints are stronger than horizontal ones. In \cite{Broughton}, if a case  in the starting case base violated the vertical constraint, then a decision for any new case cannot be forced. Instead, for us a case decided ignoring a vertical precedent is \emph{per incuriam}  and can  simply be discarded. 
Also, differently than \cite{Broughton}, we don't  assume that horizontal constraints always apply; this depends on the binding relation associated with the  model.

\cite{wyner}  extended argumentation frameworks to analyze judicial reasoning. The framework  determines which claims are justified based also on the court level, procedure type, and applicable precedents.
Future work could explore the link between our modal logic and the argumentation-based approach in \cite{wyner}.

\section*{Acknowledgements}
Antonino Rotolo was partially supported by the projects CN1 “National Centre for HPC,
Big Data and Quantum Computing” (CUP: J33C22001170001) and PE01 “Future Artificial Intelligence Research” FAIR (CUP: J33C22002830006).


 \bibliographystyle{splncs04}
 \bibliography{bibliography}

\begin{thebibliography}{10}
\providecommand{\url}[1]{\texttt{#1}}
\providecommand{\urlprefix}{URL }
\providecommand{\doi}[1]{https://doi.org/#1}

\bibitem{deonticAuth}
Araszkiewicz, M., Koszowy, M.: The structure of arguments from deontic authority and how to successfully attack them. Argumentation  \textbf{38},  171--198 (2024)

\bibitem{Ashley1990ML}
Ashley, K.D.: Modeling Legal Argument: Reasoning with Cases and Hypotheticals. MIT (1990)

\bibitem{ATKINSON2020103387}
Atkinson, K., Bench-Capon, T., Bollegala, D.: Explanation in {AI} and law: Past, present and future. Artificial Intelligence  \textbf{289},  103387 (2020)

\bibitem{capon21}
Bench-Capon, T., Atkinson, K.: Precedential constraint: {T}he role of issues. In: Proceedings of ICAIL '21. p. 12–21. ACM (2021)

\bibitem{BexP21}
Bex, F., Prakken, H.: On the relevance of algorithmic decision predictors for judicial decision making. In: Proceedings {ICAIL} '21. pp. 175--179. {ACM} (2021)

\bibitem{Blackburn_Rijke_Venema_2001}
Blackburn, P., Rijke, M.d., Venema, Y.: Modal Logic. Cambridge Tracts in Theoretical Computer Science, Cambridge University Press (2001)

\bibitem{branting}
Branting, L.K.: Reasoning with portions of precedents. In: Proceedings of ICAIL '91. ACM (1991)

\bibitem{Broughton}
Broughton, G.L.: Vertical precedents in formal models of precedential constraint. Artificial Intelligence and Law  \textbf{27}(3),  253--307 (2019)

\bibitem{Canavotto22}
Canavotto, I.: Precedential constraint derived from inconsistent case bases. In: {JURIX} 2022. {IOS} Press (2022)

\bibitem{CanovottoHierarchies}
Canavotto, I., Horty, J.: Reasoning with hierarchies of open-textured predicates. In: Proceedings of ICAIL '23. p. 52–61. ACM (2023)

\bibitem{PrecedentsClash}
Di~Florio, C., Dong, H., A., R.: When precedents clash. In: JURIX 2024, pp. 34 -- 47. IOS Press (2024)

\bibitem{Gan_Kuang_Yang_Wu_2021}
Gan, L., Kuang, K., Yang, Y., Wu, F.: Judgment prediction via injecting legal knowledge into neural networks. Proceedings of the AAAI  \textbf{35}(14),  12866--12874 (May 2021)

\bibitem{gorankoNames}
Gargov, G., Goranko, V.: Modal logic with names. Journal of Philosophical Logic  \textbf{22}(6),  607--636 (1993)

\bibitem{Horty2011RR}
Horty, J.F.: Rules and reasons in the theory of precedent. Legal theory  \textbf{17},  1--33 (2011)

\bibitem{LexisNexis}
LexisNexis: Glossary (2004), \url{https://www.lexisnexis.co.uk/legal/glossary/per-incuriam}

\bibitem{LiuLoriniJLC}
Liu, X., Lorini, E.: A unified logical framework for explanations in classifier systems. Journal of Logic and Computation  \textbf{33}(2),  485--515 (2023)

\bibitem{liu2022modelling}
Liu, X., Lorini, E., Rotolo, A., Sartor, G.: Modelling and explaining legal case-based reasoners through classifiers. In: JURIX 2022, pp. 83 -- 92. IOS Press (2022)

\bibitem{Allermuir}
{Lord Reed of Allermuir}: {D}eparting from {P}recedent: {T}he {E}xperience of the {UK Supreme Court} (2023)

\bibitem{interpreting}
MacCormick, D.N., Summers, R.S. (eds.): Interpreting Precedents: A Comparative Study. Ashgate (1997)

\bibitem{Medvedeva2020-MEDUML}
Medvedeva, M., Vols, M., Wieling, M.: Using machine learning to predict decisions of the european court of human rights. Artificial Intelligence and Law  \textbf{28}(2),  237--266 (2020)

\bibitem{canada}
Parkes, D.: {Precedent Unbound? Contemporary Approaches to Precedent in Canada}. Manitoba Law Journal  \textbf{32},  135--162 (2006)

\bibitem{Prakken2013}
Prakken, H., Wyner, A., Bench-Capon, T., Atkinson, K.: A formalization of argumentation schemes for legal case-based reasoning in aspic+. Journal of Logic and Computation  \textbf{25}(5),  1141--1166 (05 2013). \doi{10.1093/logcom/ext010}, \url{https://doi.org/10.1093/logcom/ext010}

\bibitem{Rigoni_2014}
Rigoni, A.: Common-law judicial reasoning and analogy. Legal Theory  \textbf{20}(2),  133–156 (2014)

\bibitem{CrimeAndCourtsAct}
UK, C.A.: {Crime and Courts Act 2013, Chapter 22, Part 2: Courts and Justice, Section 17, {Explanatory Notes}} (2013)

\bibitem{wyner}
Wyner, A., Bench-Capon, T.: Modelling judicial context in argumentation frameworks. Journal of Logic and Computation  \textbf{19}(6),  941--968 (2009)

\end{thebibliography}


\end{document}